\documentclass[10pt,twocolumn,letterpaper]{article}

\usepackage{cvpr}
\usepackage{times}
\usepackage{epsfig}
\usepackage{graphicx}
\usepackage{amsmath}
\usepackage{amssymb}
\usepackage{subcaption}

\usepackage[breaklinks=true,bookmarks=false]{hyperref}

\cvprfinalcopy 

\def\cvprPaperID{11} 

\ifcvprfinal\fi
\begin{document}

\title{Accurate Deep Direct Geo-Localization from Ground Imagery and Phone-Grade GPS}

\author{Shaohui Sun, Ramesh Sarukkai, Jack Kwok, Vinay Shet\\
Level5 Engineering Center, Lyft Inc\\
Palo Alto, CA, United States\\
{\tt\small \{shaohuisun, rsarukkai, jkwok, vshet\}@lyft.com}\\
}

\maketitle

\begin{abstract}
One of the most critical topics in autonomous driving or ride-sharing technology is to accurately localize vehicles in the world frame. 
In addition to common multi-view camera systems, it usually also relies on industrial grade sensors, such as LiDAR, differential GPS, high precision IMU, and etc. In this paper, we develop an approach to provide an effective solution to this problem. We propose a method to train a geo-spatial deep neural network (CNN+LSTM) to predict accurate geo-locations (latitude and longitude) using only ordinary ground imagery and low accuracy phone-grade GPS. We evaluate our approach on the open dataset released during ACM Multimedia 2017 Grand Challenge.  Having ground truth locations for training, we are able to reach nearly lane-level accuracy. We also evaluate the proposed method on our own collected images in San Francisco downtown area often described as "downtown canyon" where consumer GPS signals are extremely inaccurate. The results show the model can predict quality locations that suffice in real business applications, such as ride-sharing, only using phone-grade GPS. Unlike classic visual localization or recent PoseNet-like methods that may work well in indoor environments or small-scale outdoor environments, we avoid using a map or an SFM (structure-from-motion) model at all. More importantly, the proposed method can be scaled up without concerns over the potential failure of 3D reconstruction.
\end{abstract}

\section{Introduction}
Vision-based localization has been an active research topic for over decades. Localizing a vehicle on the road or tracking a device in an outdoor/indoor environment from an image is a fundamental problem for numerous computer vision applications. These applications include self-driving cars, Augmented Reality (AR), Virtual Reality (VR), mobile robots and etc. 

\begin{figure}[t]
\begin{center}
   \includegraphics[width=0.8\linewidth]{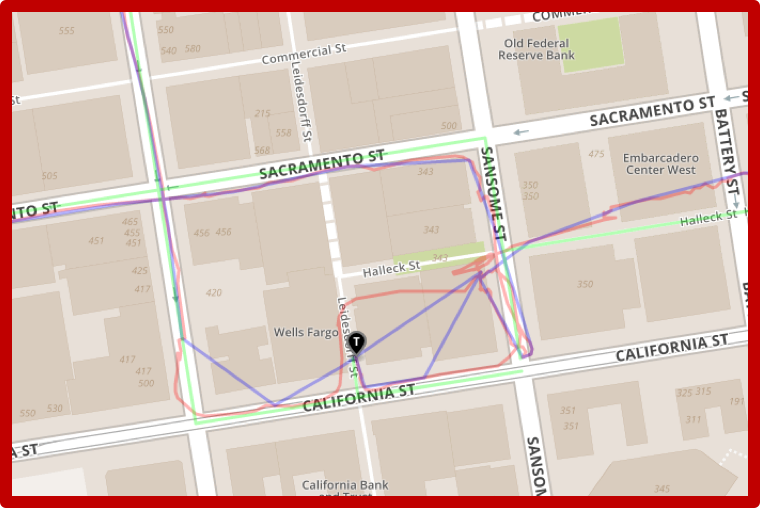}
\end{center}
   \caption{This figure shows a car ride path near San Francisco downtown where GPS signals are extremely noisy.  The red polygonal line is based on the rawly recorded GPS locations. The blue polygonal line is the filtered locations by smoothing the raw data. The green polygonal line is the corrected path (ground truth). Relying on raw GPS readings could cause irreversible damage to either self-driving navigation or for ride-sharing ETA (Estimated Time of Arrival). }
\label{fig:canyon}
\end{figure}

\subsection{Background}
Structure-from-motion (SFM) \cite{schonberger2016structure} is a relatively well-studied topic that has gain tremendous progress over the years. It takes unordered images as inputs and extracts local image features, such as SIFT, SURF, etc. and then reconstructs 3D structures of those features. Hence, given a 3D model from SFM, the problem of localizing any new image becomes a 2D-to-3D pose estimation problem. The steps are usually 1) extract 2D local features from the query image, 2) establish matches between these 2D features and the 3D points in the SFM model by computing similarities between descriptors, 3) an optimization solver such as PnP \cite{zheng2013revisiting} can take the correspondences to compute the camera pose by minimizing re-projection errors. 

Visual SLAM (Simultaneous Localization and Mapping) \cite{mur2015orb} is another popular field of research that is often adopted for most device tracking based applications. Unlike SFM which is often an offline pipeline, Visual SLAM emphasizes real-time capabilities. 

Mapping and localization are often brought up for discussion at the same time. SFM model building process is essentially a mapping process. Also, LiDAR is another great source for building 3D maps \cite{sun2013aerial}.

\subsection{What are the challenges?}
Image-based localization based on SFM or Visual SLAM always requires a decently reconstructed 3D model to start with. It can become fairly challenging especially when the scene is not favored by SFM or Visual SLAM. Commonly known factors, such as inconsistent illumination, motion blur, texture-less surfaces, lack of overlaps between images, can easily cause failures by using these local feature dependent localization approaches. Being able to estimate 6DoF pose accurately is absolutely crucial in a small environment especially when the application needs to precisely place virtual objects on the real physical surface. However, due to limitations mentioned above, a number of difficulties remain. Aside from everything else, getting a complete and decent SFM model is never an easy nor worry-free task. PoseNet \cite{kendall2015posenet} tries to solve this task by formulating it as a machine learning problem and shows promising results but with questionable quality. Without exception, its training process requires good SFM models ready for use.  Above all, 6DoF localization only makes sense when there is a so-call HD map in place. That being said, HD mapping still remains an open topic for both academia and industries to research and discuss. An industrial standard does not even exist yet.

In most ride-share or vehicle navigation businesses, GPS is the only source for localization with the assistance of some standard map service (eg. Google Maps, Apple Maps). Getting accurate latitude and longitude values are critical to the services provided by these businesses. For instance, when a customer requests a ride through any ride-share application (eg. Lyft, Uber), ETA (estimated time of arrival) which is directly tied to the quality of user experience and the fairness of pricing is largely determined by latitude and longitude measurements. As is commonly known, phone-grade GPS receivers are easily affected by a variety of factors, such as atmospheric uncertainty, building blockage, multi-path bounced signals, satellite biases, etc.  In Fig.\ref{fig:canyon}, it shows a car ride recorded in some urban area that fits the description as urban canyon where GPS signals can be occasionally entirely out of touch.  The green path is the actual ride path. The red one is the path recorded by GPS readings. The blue one is the filtered version after some unsophisticated smoothing. As we can see, neither raw readings nor smoothed ones can actually represent the real ride path. Aside from low accuracy, low frequency about phone-grade GPS also prevents on-road navigation from being precise. How to overcome this has become increasingly important to real-world applications.

To summarize, the main challenges are:

(1) Computing a map from SFM or SLAM is not easy. It can fail surprisingly more often than expected due to multiple factors relative to image quality, scene content characteristics, etc. 

(2) Assuming a good map in place, image-based localization by matching 2D-3D point correspondences can also easily fail due to same limitations mentioned in (1).

(3) Low-priced phone-grade GPS is noisy. Unfortunately, it is often the only source used for localization in many real-world applications, eg. ride-sharing, car navigation.

(4) High-end GPS equipment is extremely expensive and is not practical to be installed on a large fleet of vehicles.

\subsection{Our contribution}
In this work, we propose a framework to directly infer a more accurate GPS location, [$Lat, Lon$], from input imagery.  The overall idea is to learn to predict the distance between the noisy GPS location and the true location. Using the trained knowledge, we can compensate the error of raw GPS data under the reality that there is no knowledge about where the true location is. To do so, we take the image and the corresponding raw noisy GPS reading and leverage a pre-trained Convolutional Neural Network to learn suitable feature representations for our particular localization purpose, and then we make use of Long-Short Term Memory (LSTM) units \cite{Hochreiter:1997} on the final FC layer out of CNN. We train the model with real ground truth for each recorded location. The evaluation of an open dataset shows that we can achieve near lane-level accuracy using an image and noisy GPS only. We also trained a model without any ground truth but with sparsely hand-picked ground control points. In problematic areas like urban canyons, the model can predict the location of an input image with or without raw GPS. Based on observation, the model also behaves as a much more sophisticated smoothing filter that tries to correct wrong GPS readings. 

To summarize, our major contributions are listed as below: 

(1) We demonstrated the power of CNN + LSTM architecture to regress near lane level accurate geo-locations for vehicle navigations when raw GPS is available but noisy;

(2) We provide a solution that relies upon no HD maps or SFM models. It applies to both training and inferring. We only use images and raw GPS data to predict more accurate geo-location.

(3) The proposed approach can predict an accurate location without any raw GPS. It is very helpful when losing GPS signal in downtown canyons. We also show that the trained model can function as a filter to smooth noisy GPS data. 

\section{Related work}
\subsection{Visual inertial localization}
The traditional ways to approach the localization problem are relying on structure-based techniques. It uses a image-derived 3D model usually obtained from Structure-From-Motion (SFM) as a map. 6DoF pose estimation of a query image is done by matching point features found in both the 2D image and the 3D point cloud. Recent advances in SFM \cite{schonberger2016structure} allow to reconstruct large scenes and hence provide a better model for image-based localization. The main challenge here is the search complexity that could grow exponentially high as the model size get increasingly larger. There are some works in prioritized matching \cite{li2010location} \cite{sattler2017efficient}. They first consider features more likely to be matched and terminate the search process as soon as enough matches have been found.

Visual inertial camera tracking or re-localization has gained significant attention, especially in the field of Augmented Reality. In this case, the camera and the inertial sensors (IMU) complement each other in a joint optimization framework. Most of VI localization methods perform well in indoor environments. For outdoor scenarios, \cite{Lynen2015GetOO} addresses how they tackle the complexity of localization against a large map. They demonstrate that large-scale, real-time 6DoF localization can be performed on mobile platforms with limited resources without the use of a server.

Overall, the run-time of traditional localization approaches is determined by the number of 2D and 3D features that are engaged in optimization. Therefore, scalability is put in question constantly. In addition, local feature based methods do not perform in numerous situations due to the common challenges in image processing. This further encourages the exploration of using an alternative approach based on deep learning.

\subsection{Conventional machine learning based localization}
There is a good amount of work in location recognition using conventional machine learning techniques.  In \cite{gronat2013learning}, the authors addressed the challenges when dealing with visual place recognition. Changes in viewpoint, imaging conditions and the large size of geotagged image database make this task very challenging. Bag-of-words methods are favorable in this category. In \cite{Cao_2013_CVPR}, the authors choose to represent the database as a graph and show the rich information embedded in a graph can improve a bag-of-words based location recognition method.

However, this type of location recognition usually only produces coarse location information. It is certainly useful in automated image geotagging, while it is not accurate enough for navigation purposes.  

\subsection{Deep learning based localization}
Deep learning techniques, especially convolutional neural networks (CNN) have been successfully applied to most tasks in computer vision. A great number of tasks are already beyond image classification and object detection. Deep learning has driven the machine learning the focus from hard-core feature engineering to high volume data manipulation. How to improve performance has shifted from algorithm-driven to data-driven. However, the need for large datasets for training is also a drawback for deep learning. Hence, a common solution is called transfer learning. Fine-tuning modified pre-trained networks on a much smaller dataset for a more specific domain-related task has become quite essential in most computer vision research. Long Short-Term Memory (LSTM) \cite{Hochreiter:1997} is a type of Recurrent Neural Network (RNN) that is designed to accumulate or abandon relevant contextual information in hidden states. In recent years, CNN and LSTM have been placed in one unified framework for tasks such as various video analysis problems, human action analysis, etc. CNNs are good at reducing variations in frequency, while LSTMs are good at temporal modeling \cite{sainath2015convolutional}.

In \cite{kendall2015posenet}, the authors present a robust and real-time monocular 6DoF re-localization system which is known as PoseNet. It introduces an end-to-end regression solution with no need for additional engineering or graph-based optimization. In \cite{kendall2016modelling}, an extension to PoseNet evaluates the CNN with a fraction of its neurons randomly disabled. It results in different pose estimations that can model the uncertainty of the poses. The problem of PoseNet is it is relatively inaccurate. \cite{Walch_2017_ICCV} proposes a new CNN+LSTM architecture for pose regression and provides an extensive quantitative comparison of CNN-based and SIFT-based localization methods.

In this paper, we show how CNN + LSTM architecture is capable of predicting very accurate location for navigation purposes.

\begin{figure*}
     \centering
     \includegraphics[width=0.9\textwidth]{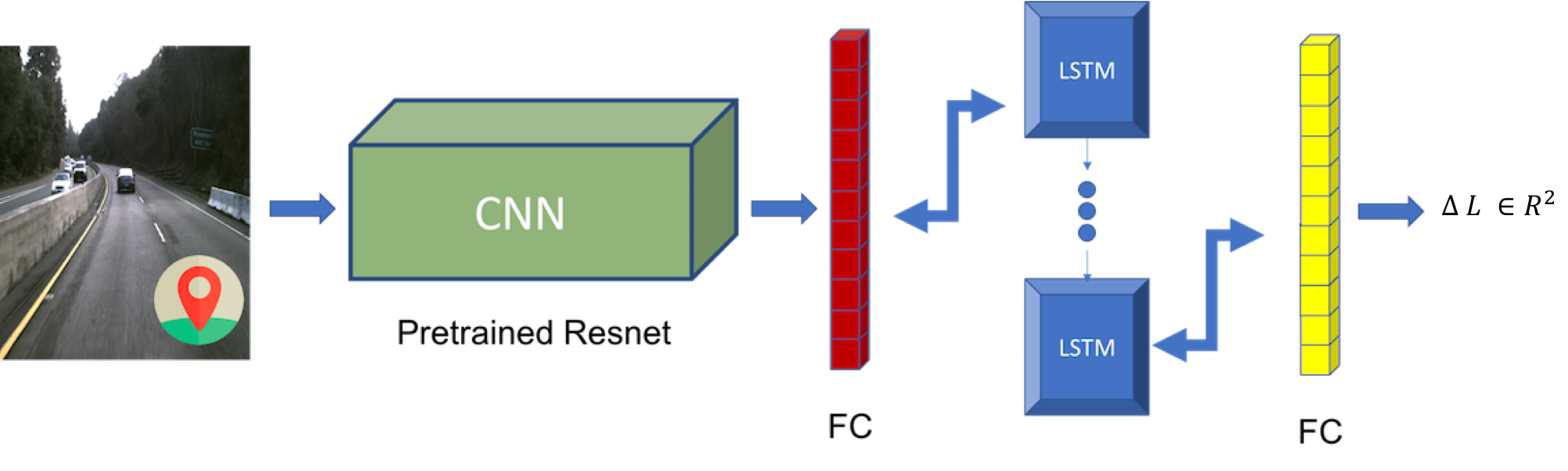}
     \caption{Architecture of the proposed location regression CNN LSTM network}
\label{fig:architecture}
\end{figure*}

\section{Choice of architecture}

The main goal of this work is to prove the state-of-art deep learning technology can bring a scalable solution to geo-localization that is needed in either ride-sharing or autonomous driving industries. PoseNet \cite{kendall2015posenet} simply adopted GoogleNet with a few necessary modifications due to the regression purpose. In \cite{walch2016image}, an LSTM layer is introduced in addition to the modified GoogleNet in \cite{kendall2015posenet} even though the input is not a typical sequential data, and it improves the overall performance. Therefore, we also adopted a CNN + LSTM architecture with modifications. In \cite{walch2016image}, reshaping the input feature vector to LSTM and breaking it into smaller parts are done to increase the regression accuracy. In our experiments, reshaping the vector actually downgrades the performance a little. Hence, we choose not to reshape the input feature vector in order to provide the best performance (prediction error on locations).

\section{Deep direct localization}
In this session, we develop our approach to learn to regress accurate geo-locations, normally represented as [latitude, longitude] in most navigation scenarios, directly from ground imagery that could be taken from a in-vehicle dash camera or phone camera mounted behind the windshield and the raw geo-location recorded by a phone-grade GPS receiver usually at a very low frequency (1 Hz). In practice, it is extremely challenging to infer absolute locations from images. Our main goal is to train a CNN + LSTM network to learn a mapping function from an image to a difference location relative to the 'true' location or the hand-picked ground control points, $f(I) = \Delta l$, where $f(\cdot)$ is the neural network, $\Delta l \in \mathbb{R}^{2}$ . Each $\Delta l$ comprises of $\Delta lat$ and $\Delta lon$. We adopt an architecture that is similar to the one in \cite{Walch_2017_ICCV}. Our architecture is depicted in Figure.\ref{fig:architecture}. All hyperparameters used for the experiments are detailed in Section \ref{experiments}. The Smooth L1 loss function is chosen for the sake of stability. 

\begin{equation}
L_{loc}(\Delta l_{i} - \Delta \tilde l_{i}) = Smooth_{L_{1}}(\Delta l_{i} - \Delta \tilde l_{i})
\end{equation}

which

\begin{equation}
Smooth_{L_{1}}(\Delta l_{i} - \Delta \tilde l_{i}) = 
\begin{cases}
    0.5 x^{2},& \text{if } |x| < 1\\
    |x| - 0.5,              & \text{otherwise}
\end{cases}
\end{equation}

\subsection{CNN feature extraction}
It is a common practice not to train a convolutional neural network from scratch. Training from scratch usually requires a really large dataset which brings a huge cost in numerous ways.  Unlike classification problems which demand at least one sample for each label, the output space for regression problems is continuous and infinite in theory. Therefore, transfer learning is a highly effective approach. We take advantage of the pre-trained state-of-art classification network ResNet \cite{he2016deep} and modify the last fully connected layer to output a $n$-dimensional vector (Figure. \ref{fig:architecture}). One can directly reduce the dimension of the FC to be the dimension of the desired output. Intuitively, we can define $n$ to be $2$, so this reduced $2$-element vector is the final regressed difference location that we target for. However, from our empirical experience, it produces poor results. Hence, this $n$-dimensional vector is the input to the following recurrent neural network and can be practically perceived as a concise representation of the original image to be localized.

\subsection{Location regression with LSTMs}
Long Short-Term Memory (LSTM) units are typically applied to sequential data that are embedded with rich temporal information, such as natural language processing, video action unit analysis. But, the capability of LSTM is not limited to only temporal sequences. In our case, the $n$-dimensional vector from the ResNet CNN can be regarded as a sequence. Two or more LTSM layers can be inserted after the FC from the CNN. No special vector reshaping is required. Most of the time, we choose 2 LSTM units which can perform well enough, and no major benefit gain even if using more LSTM unites.

\begin{figure}
     \centering
     \includegraphics[width=0.5\textwidth]{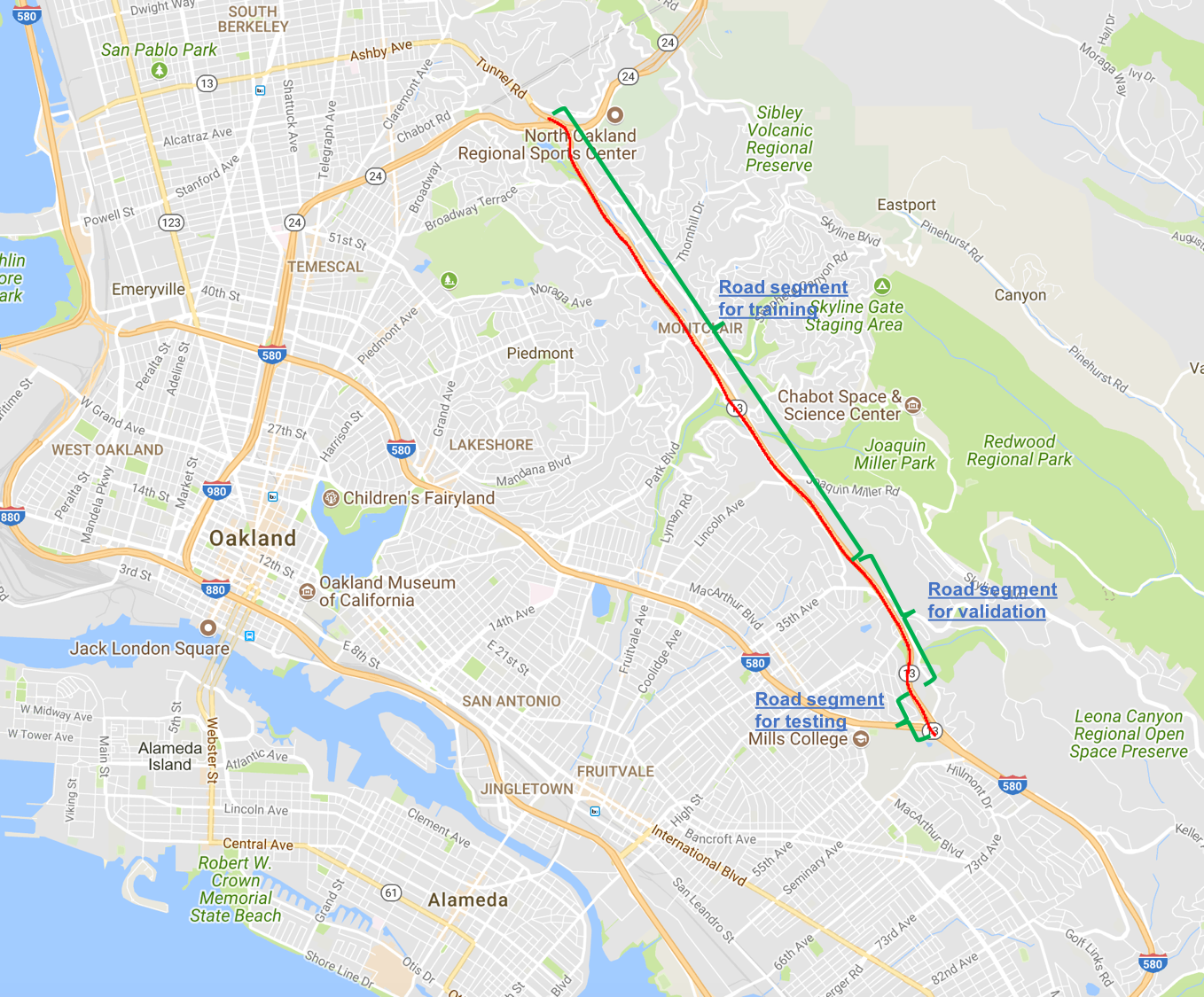}
     \caption{The red line shows the trajectory covered by the dataset from "Lane Level Localization on a 3D Map". Three segments are proportionally selected along the path for training, validation and testing.}
\label{fig:datamap}
\end{figure}

\begin{figure}
     \centering
     \includegraphics[width=0.5\textwidth]{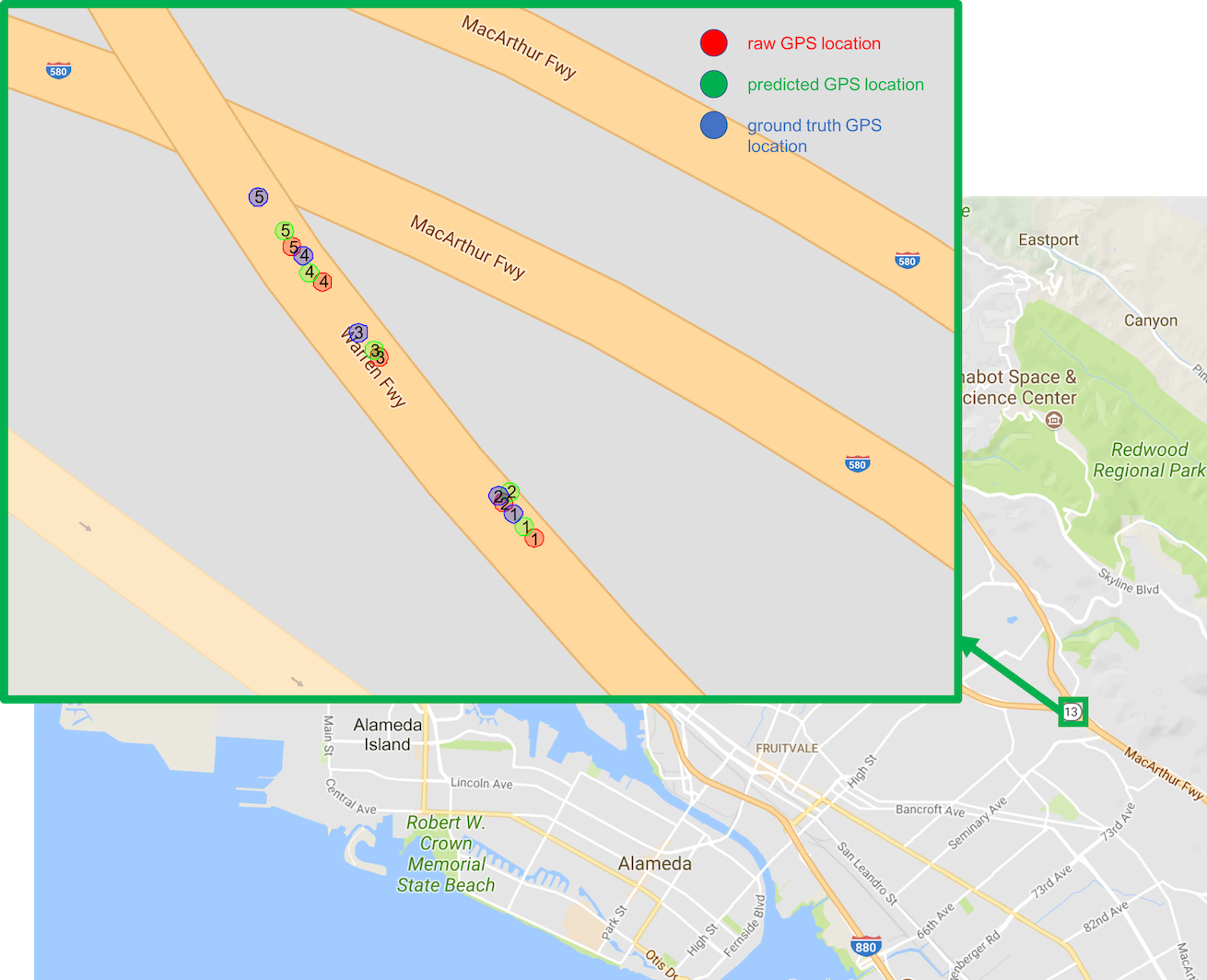}
     \caption{Five randomly picked points shown on the real map. Raw points colored in red. Predicted points are colored in green. True points are colored in blue.}
\label{fig:acm_results}
\end{figure}

\begin{figure}
     \centering
     \includegraphics[width=0.5\textwidth]{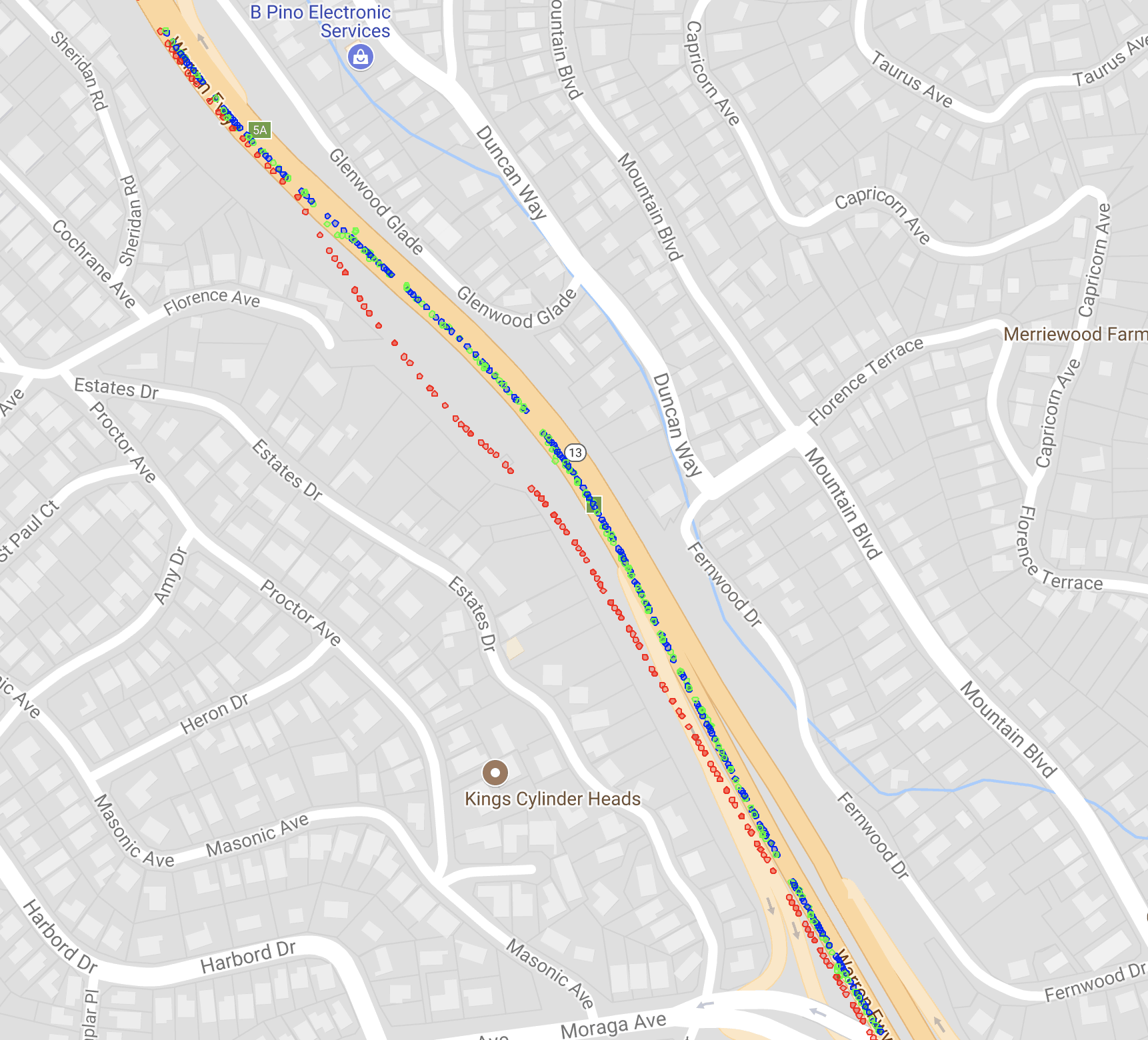}
     \caption{A better look on results of ACM dataset. Raw points colored in red. Predicted points are colored in green. True points are colored in blue. In this segment, raw GPS are very off.}
\label{fig:acm_results2}
\end{figure}

\begin{figure*}
     \centering
     \includegraphics[width=0.85\textwidth]{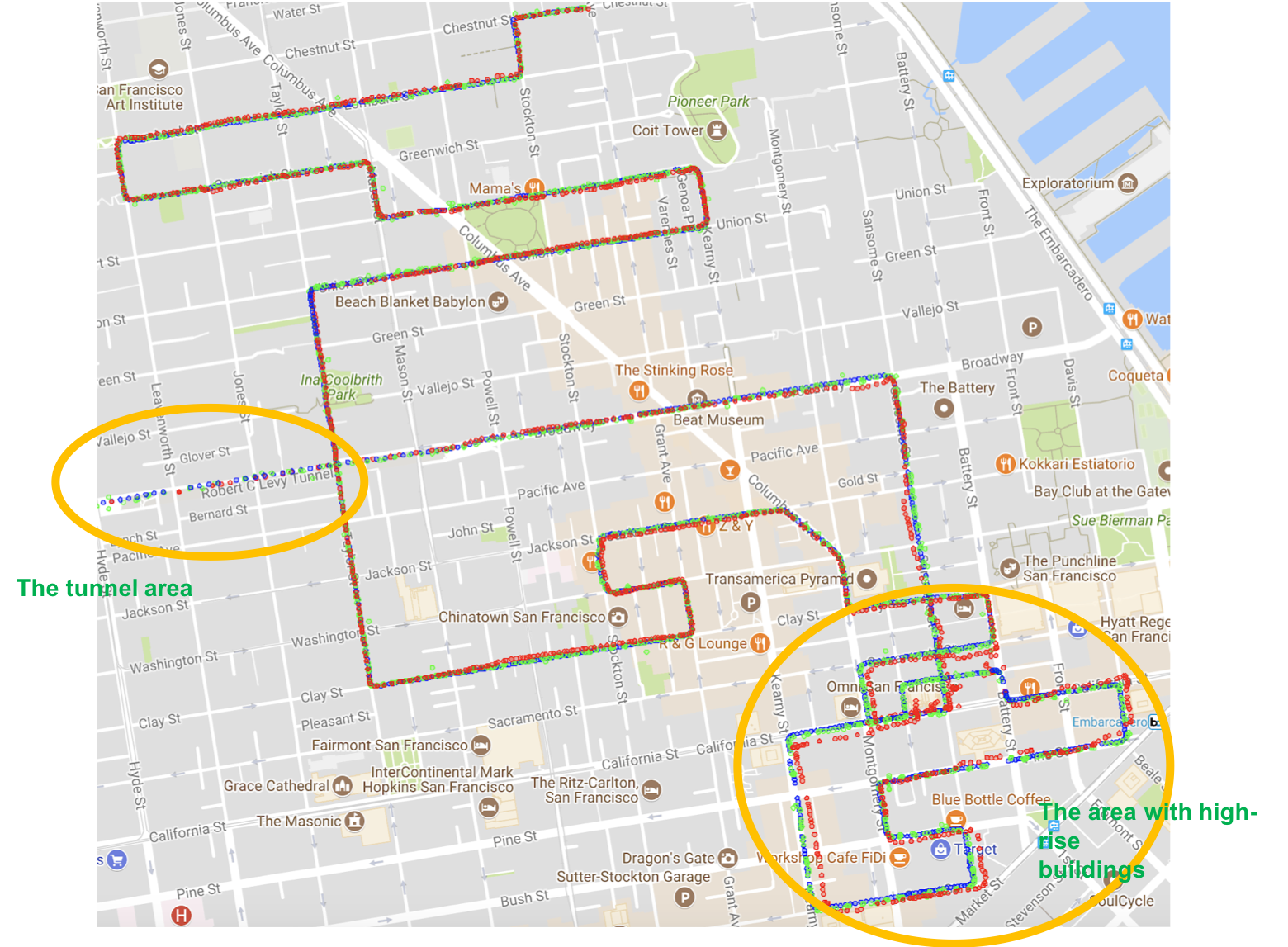}
     \caption{Results from San Francisco data shown on the real map. Raw points colored in red. Predicted points are colored in green. True points are colored in blue.}
\label{fig:sf_results_big}
\end{figure*}

\begin{figure}
     \centering
     \includegraphics[width=0.5\textwidth]{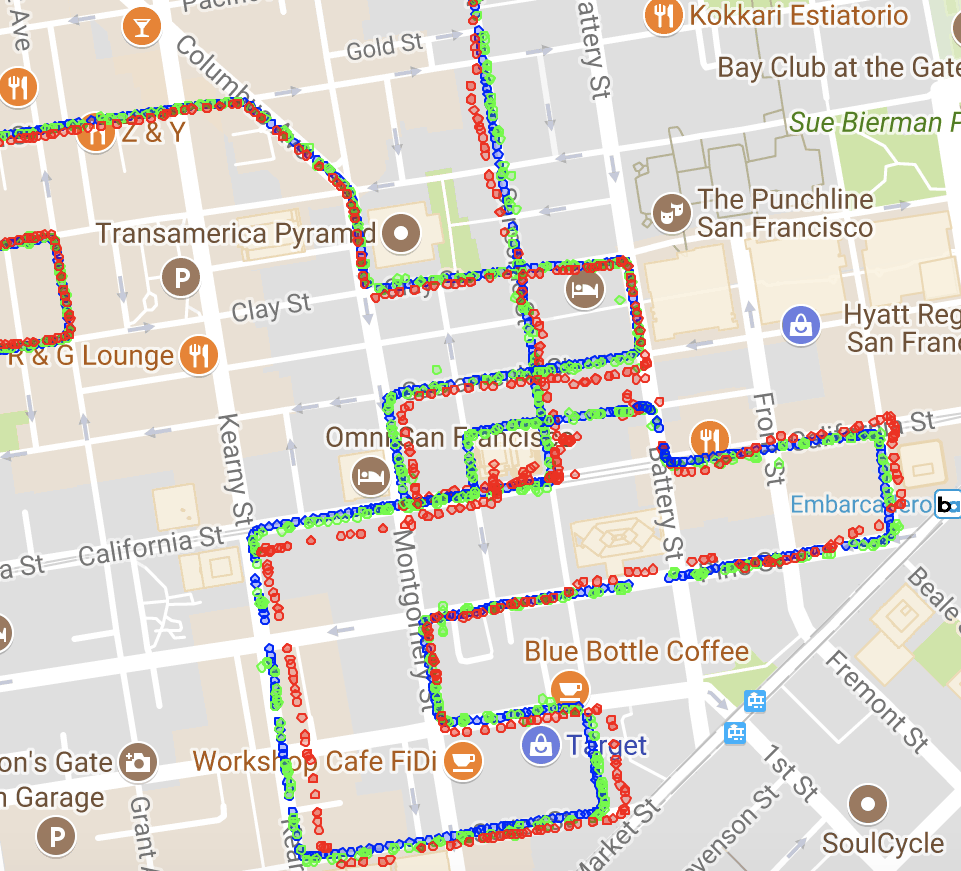}
     \caption{A closer look on the financial district in SF. Raw points colored in red. Predicted points are colored in green. True points are colored in blue.}
\label{fig:sf_results}
\end{figure}

\section{Experiments} \label{experiments}
\subsection{Experiment setup}
We conduct all experiments using PyTorch on a single GPU machine equipped with one Geforce GTX 1080 card. We initialize part of parameters from pre-trained ResNet model and randomly initialize the remaining weights. All input images are resized to 224 x 224 pixel. Radom image cropping is used during training. SGD is chosen to be the optimizer with the learning rate at 0.045. Random shuffling is performed for each batch. We use small batch size such as 8. 

\begin{figure}
\centering
\begin{subfigure}{.49\columnwidth}
  \centering
  \includegraphics[width=.8\linewidth]{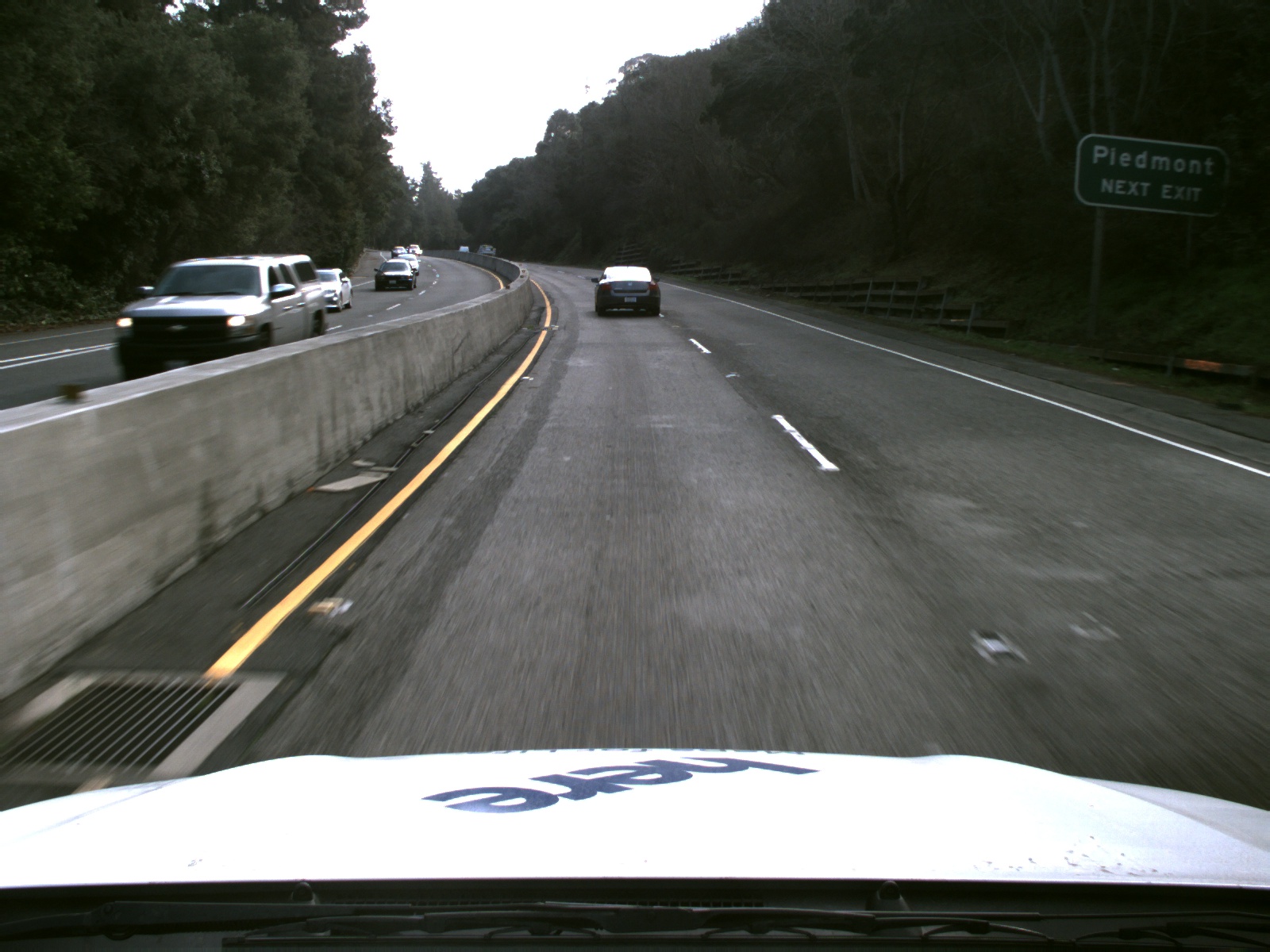}
\end{subfigure}
\begin{subfigure}{.49\columnwidth}
  \centering
  \includegraphics[width=.8\linewidth]{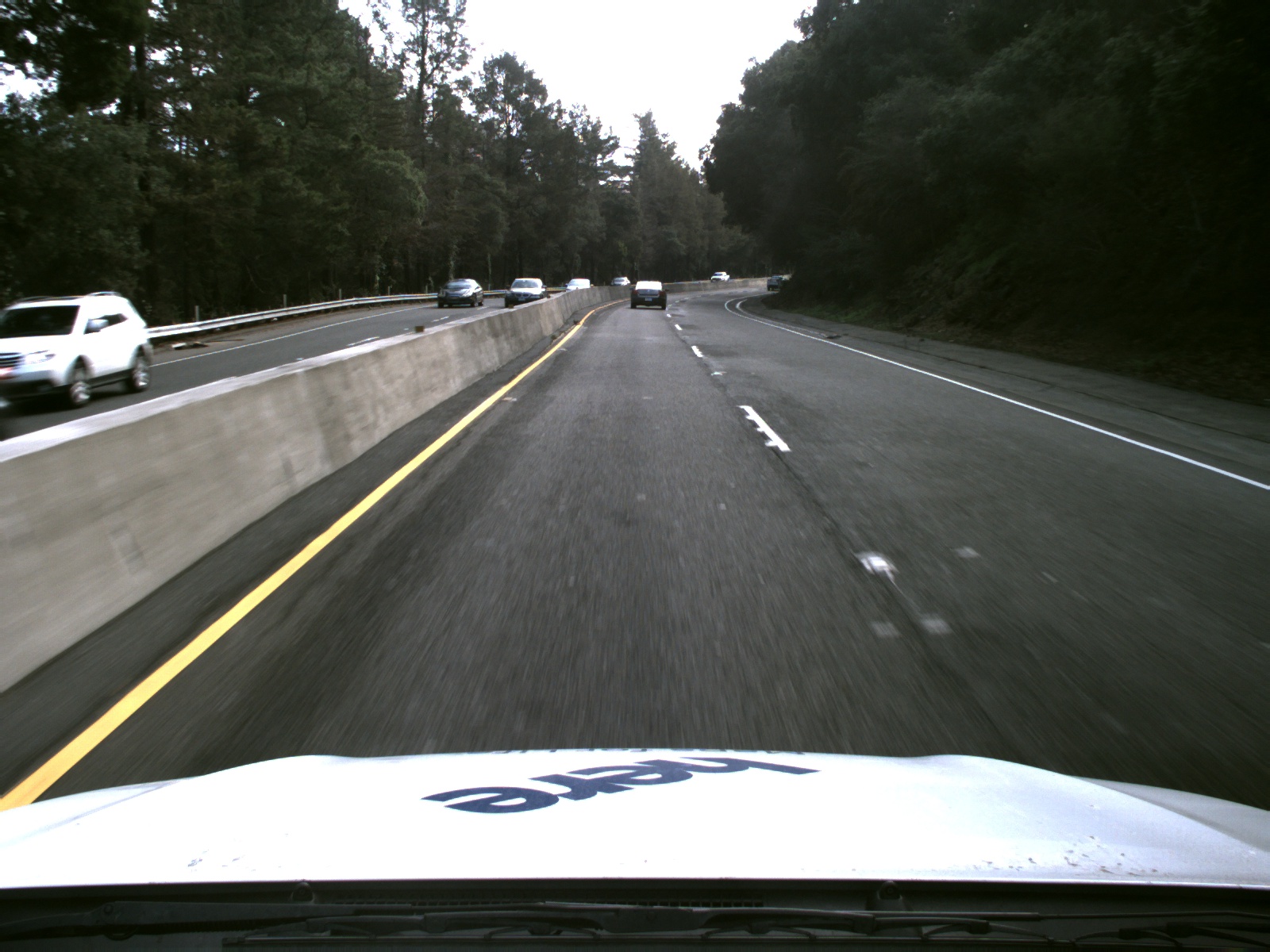}
\end{subfigure}
\caption{Some sample images from the ACM dataset}
\label{fig:test}
\end{figure}

\subsection{Training with high precision GPS}
We first choose to use the dataset released as part of the ACMMM 2017 grand challenge "Lane Level Localization on a 3D Map" \cite{andi2017}. This is the only dataset we can find publicly that satisfies our specific requirements that both true phone-grade GPS and industrial-grade GPS are in place. The dataset contains around 3000 images (sample images can be seen in Figure.\ref{fig:test}) acquired with a commercial webcam at 10 Hz, a set of consumer phone grade GPS points synchronized with the image timestamp, 3D map information (eg. road and lane boundaries, traffic sign location, occupancy grid in voxels), and camera intrinsic parameters. The data covers over 20km. Ground truth GPS points acquired from a survey-grade GPS device are also given for training and testing purpose. The whole trajectory is shown in Figure. \ref{fig:datamap}. We divide the whole trajectory into three segments for training, validation and testing purposes respectively (also shown in Figure. \ref{fig:datamap}). As we address in the beginning, we do not rely on 3D HD map. Hence, we only utilize a subset of the whole dataset which includes images, phone-grade GPS points, and survey-grade GPS points. A preprocessing step is first done to synchronize the frames based timestamps. The measurement error range of phone-grade GPS points is from 0.37419 to 61.7118 meters. The mean error is 9.8772 meters, and the standard deviation is 11.7547 meters.

\begin{figure}
     \centering
     \includegraphics[width=0.5\textwidth]{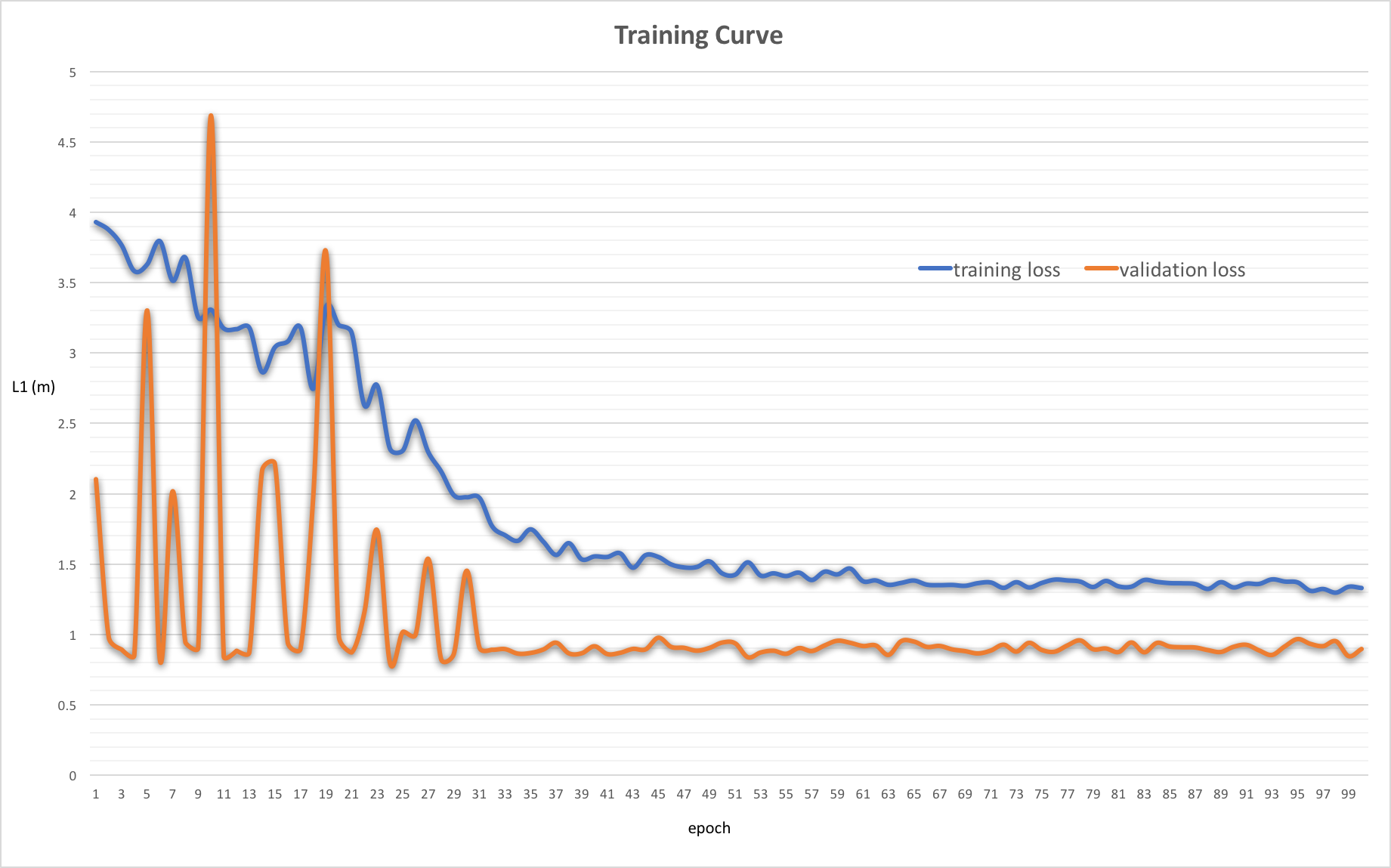}
     \caption{Traning curve over the ACM dataset}
\label{fig:acm_training_curve}
\end{figure}

Each data sample contains the image, the raw phone grade latitude and longitude values, and the distance between the raw GPS value and the ground truth. The training curve can be seen in Figure. \ref{fig:acm_training_curve}. The evaluation metric used in this experiment here is the $L2$ distance in meters between the true location and the predicted location. Note, it is not convenient to directly use latitude and longitude to compute the small distance between two points. Hence, UTM coordinates are actually used to compute the distance. The conversion between UTM coordinates and Lat-Lons is a necessary step here.

In Figure.\ref{fig:acm_results} and Figure.\ref{fig:acm_results2}, one can visually examine some location points, and their predicted points and true points respectively. Again, it demonstrates visually that accuracy of the GPS measurements is improved. Please see Table.\ref{table1} for actual values of our prediction error.

\begin{table}[h]
\center
\begin{tabular}{|c |}
\hline
 Our prediction error \\
\hline
mean: 2.47 std: 1.58 \\
\hline
\end{tabular}
\caption{Prediction errors on ACM dataset (unit: meter)}
\label{table1}
\end{table}

We also test on our internally collected dataset covering a big portion of San Francisco using an in-house built application on Android phones. In this dataset (Figure.\ref{fig:test2}), challenging scenarios,such as urban canyons and tunnels, are covered. Please see Figure.\ref{fig:sf_results_big} and Figure.\ref{fig:sf_results} for the results. A demo video is provided as the supplemental material from this dataset.

In the United States, the Interstate Highway standards for the U.S. Interstate Highway System uses a 12-foot (3.7 m) standard for lane width. With the level of accuracy shown above (2 m), we can confidently claim it gets near lane-level accuracy.

\subsection{Training without high precision GPS}

\begin{figure}
\centering
\begin{subfigure}{.49\columnwidth}
  \centering
  \includegraphics[width=0.6\linewidth]{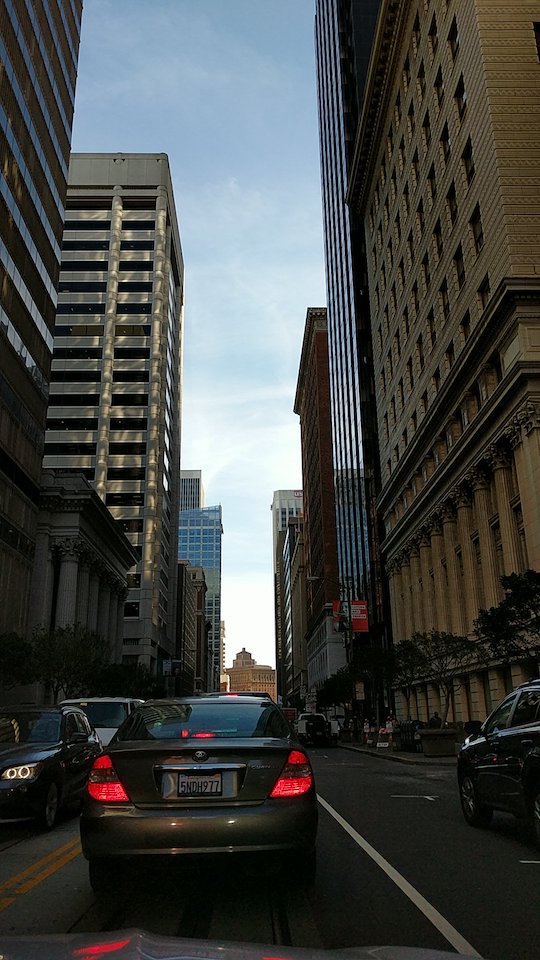}
\end{subfigure}
\begin{subfigure}{.49\columnwidth}
  \centering
  \includegraphics[width=0.6\linewidth]{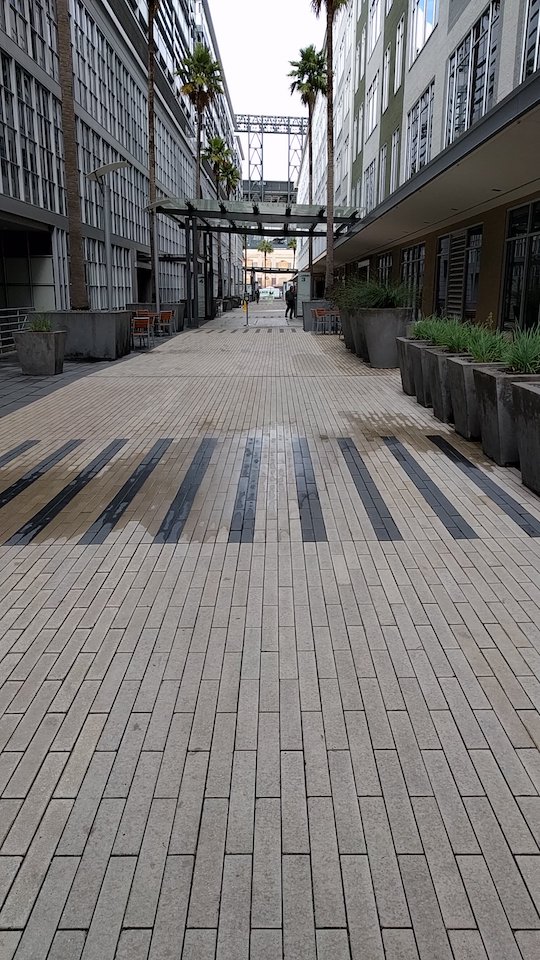}
\end{subfigure}
\caption{Some sample images from the SF dataset and the Courtyard dataset}
\label{fig:test2}
\end{figure}

\begin{figure}
     \centering
     \includegraphics[width=0.5\textwidth]{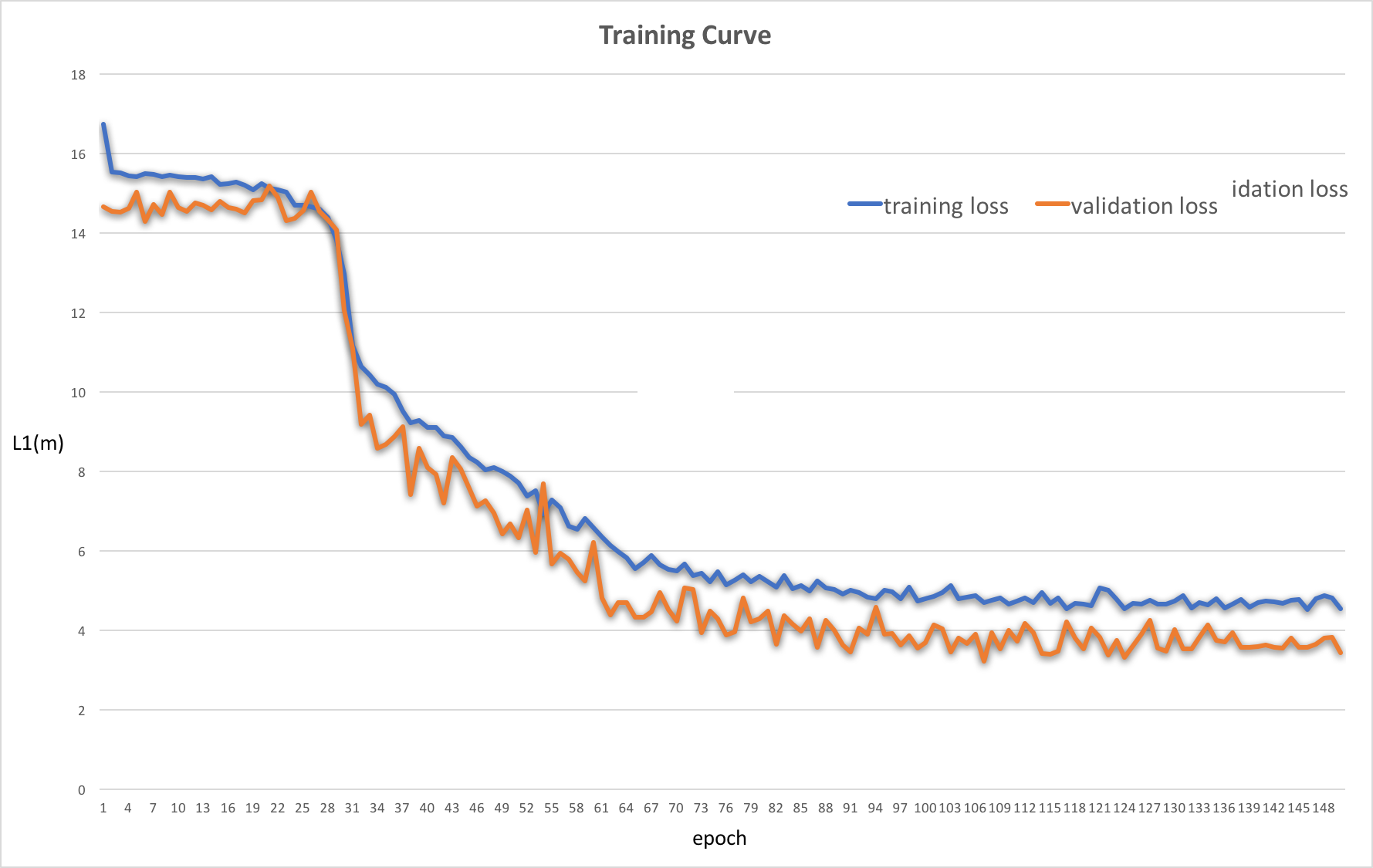}
     \caption{Training curve over the Courtyard dataset}
\label{fig:courtyard_training_curve}
\end{figure}

\begin{figure}
     \centering
     \includegraphics[width=0.5\textwidth]{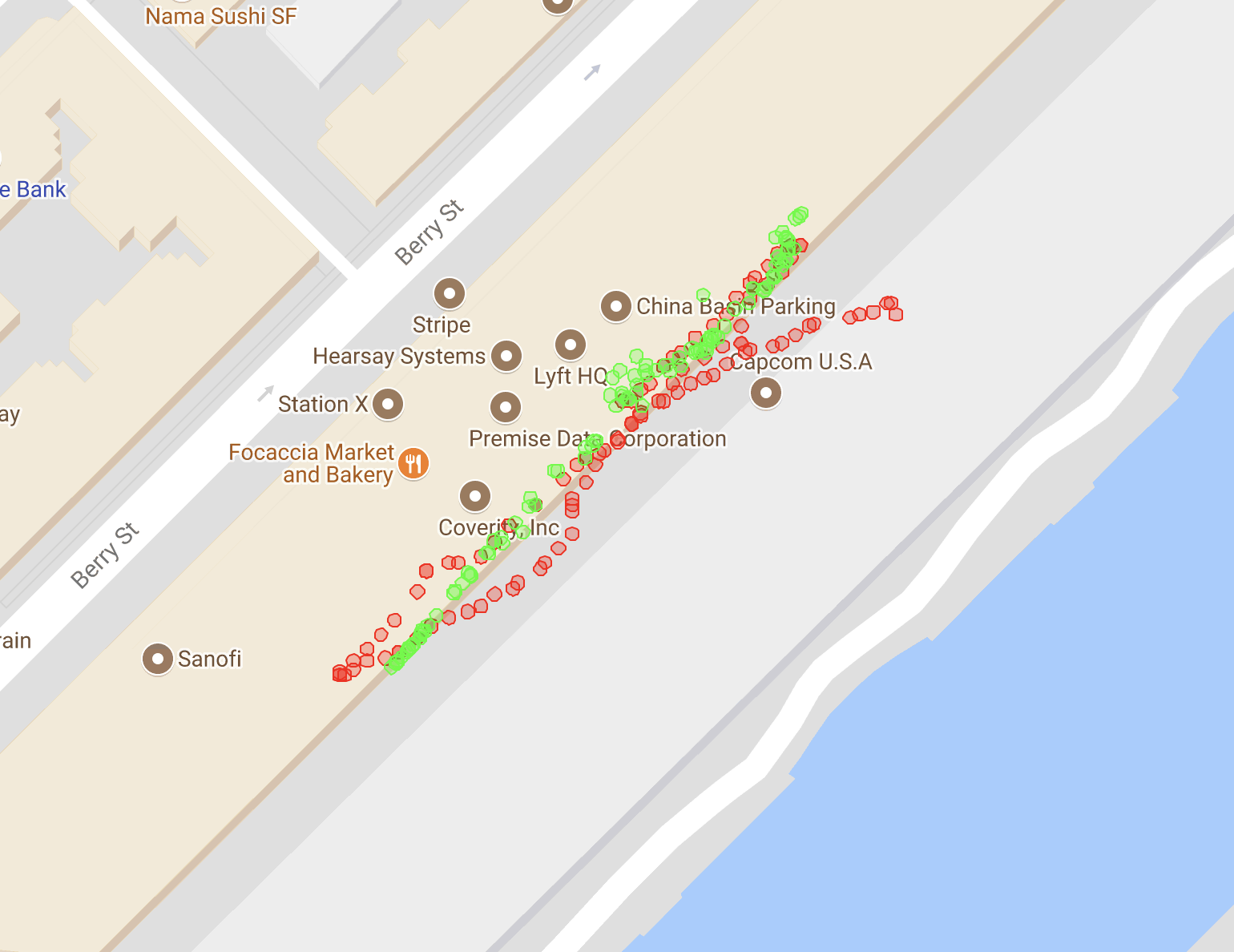}
     \caption{Courtyard dataset: red dots are raw GPSs; green dots are predicted GPSs}
\label{fig:courtyard_results}
\end{figure}

We collected another dataset using the same Android App at some courtyard in between two moderately high buildings somewhere around Downtown San Francisco (Figure.\ref{fig:test2}). The facades of the buildings negatively affect the GPS signals. A phone-grade GPS will not be able to provide accurate reading and can sometimes even be unexpectedly confused by WIFI signals from inside the buildings. Before starting the collection, we have to make sure the WIFI receiver on the device is switched off. The image collection can at 5 to 10 Hz, while the GPS is recorded at 1 Hz. We walk along the path back and forth many times and also on different days. We roughly walk following a straight line during all data collections. In this dataset, we are not able to mark where the true locations are. Nonetheless, we can still use the proposed framework to train by manually picking a known nearby location from the map as the ground control point. The model then predicts the distance between the image location and this ground control point. Each data sample in this experiment contains the image, the raw GPS, and the distance between the GPS point and the ground control point. The evaluation metric is the same as the previous experiment. The training curve can be seen in Figure.\ref{fig:courtyard_training_curve}. As a reminder, we want to emphasize that inferring is done with only images and the known ground control point. No raw GPS was used for inference. From Figure.\ref{fig:courtyard_results}, we can tell all predicted points are closer to a center line while the raw GPS behave in a more arbitrary way due to the impact of the noise.

\subsection{KITTI dataset}
We further test our method on the KITTI dataset. Please refer to Table\ref{table2} for results on three sequences covering a relatively large area.  Note, the KITTI dataset does not provide phone-grade GPS. So, we introduced simulated errors to the original GPS data to get the noisy GPS needed.  For reference, ORB-SLAM2 in stereo mode can only achieve up to 1.15 meters on KITTI dataset. MLM-SFM can only achieve up to 2.54 meters.

\begin{table}[h]
\center
\begin{tabular}{|c | c | c | }
\hline
Sequence name  & Raw error & Our prediction error \\
\hline
2011 10 03 drive 0027 & 8.58 & 2.05 \\
2011 09 29 drive 0071 & 8.94 & 1.56 \\
2011 10 03 drive 0042 & 8.59 & 1.53  \\
\hline
\end{tabular}
\caption{Prediction results on KITTI dataset (unit: meter)}
\label{table2}
\end{table}

\section{Conclusion}
In this paper, we address the challenge of accurate localization from imagery for ride-share or car navigation businesses. We use a hybrid deep learning architecture that combines a CNN with LSTM units to regress geo-locations directly. We don't rely on any pre-computed HD map or SFM model during either training or inferring. The trained model is able to predict near lane-level locations from imagery and noisy raw GPS, and it can also infer accurate locations without GPS as prior. Furthermore, this is the first work where deep learning is applied to the problem of directly localizing to GPS Lat-Lon applied to real-world ride-sharing and navigation problems.

Future work will look at expanding to larger datasets. However, the challenge is getting much larger data sets than the ACM dataset. Therefore, making a larger public dataset for direct geo-location learning from imagery is to be on our agenda. In the meanwhile, we need to understand better how a localization network actually behaves at different stages so that we can impose more control and increase the performance.

{\small
\bibliographystyle{ieee}
\bibliography{egbib}
}

\end{document}